\pdfoutput=1
\documentclass[11pt]{article}

\usepackage[final]{acl}

\usepackage{listings}
\usepackage{times}
\usepackage{latexsym}
\usepackage{tabularx}
\usepackage{xcolor}
\usepackage[T1]{fontenc}
\usepackage[utf8]{inputenc}

\usepackage{microtype}
\usepackage{amsmath}
\usepackage{amssymb}
\usepackage{inconsolata}
\usepackage{hyperref}
\usepackage{multirow}
\usepackage{bbding}
\usepackage{booktabs}
\usepackage{graphicx}

\title{Learning to Plan for Retrieval-Augmented Large Language Models from Knowledge Graphs}

\author{
    Junjie Wang\textsuperscript{1,2,5}\footnotemark[1],
    Mingyang Chen\textsuperscript{3}\thanks{~~Equal contribution.},
    Binbin Hu\textsuperscript{2,5}, 
    {Dan Yang}\textsuperscript{2,5},
    {Ziqi Liu}\textsuperscript{2,5},\\
    \textbf{Yue Shen\textsuperscript{2,5}},
    \textbf{Peng Wei\textsuperscript{2,5}},
    \textbf{Zhiqiang Zhang\textsuperscript{2,5}},
    \textbf{Jinjie Gu\textsuperscript{2,5}},
    \textbf{Jun Zhou\textsuperscript{2,5}},\\
    \textbf{Jeff Z. Pan\textsuperscript{4}},
    \textbf{Wen Zhang\textsuperscript{1,5}\footnotemark[2]},
    \textbf{Huajun Chen\textsuperscript{1,5}}\thanks{~~Corresponding authors.}\\
    \textsuperscript{1}Zhejiang University,
    \textsuperscript{2}Ant Group,
    \textsuperscript{3}Baichuan Inc.,
    \textsuperscript{4}The University of Edinburgh \\
    \textsuperscript{5}Zhejiang University - Ant Group Joint Laboratory of Knowledge Graph \\
    \texttt{
    \{wangjj2018,zhang.wen,huajunsir\}@zju.edu.cn, chenmingyang@baichuan-inc.com
    }\\
    \url{http://knowledge-representation.org/j.z.pan/}\\
  \url{https://github.com/zjukg/LPKG}
}
\begin{document}
\maketitle
\begin{abstract}
Improving the performance of large language models (LLMs) in complex question-answering (QA) scenarios has always been a research focal point. Recent studies have attempted to enhance LLMs' performance by combining step-wise planning with external retrieval. While effective for advanced models like GPT-3.5, smaller LLMs face challenges in decomposing complex questions, necessitating supervised fine-tuning. Previous work has relied on manual annotation and knowledge distillation from teacher LLMs, which are time-consuming and not accurate enough. In this paper, we introduce a novel framework for enhancing LLMs' planning capabilities by using planning data derived from knowledge graphs (KGs). LLMs fine-tuned with this data have improved planning capabilities, better equipping them to handle complex QA tasks that involve retrieval. Evaluations on multiple datasets, including our newly proposed benchmark, highlight the effectiveness of our framework and the benefits of KG-derived planning data.

\end{abstract}

\section{Introduction}

% 1. Methods for improving LLM performance in complex QA: a) prompting like CoT, ToT; b) RAG,Iter-RAG; c) Combination, like ReAct,Self-Ask.
% For c), smaller LLMs is not good enough, thus need sft.

% 2. Challenge: The acquisition of tuning data. Human annotation is time-consuming and labor-intensive; Distillation is not always correct.

% 3. Introduction of KG pattern, motivation. 

% 4. Introduction of our benchmark(?)

% 5. Contribution:a) A new way for acquiring planning data from KG patterns; b) Achieve GPT-3.5 level planning ability, and outperform normal distillation method; c) A more challenging benchmark.

% Figure 1 KG pattern and Multihop QA

The past few years have witnessed significant innovations in LLMs \cite{InstructGPT,llama2,PaLM,llama3modelcard}. While LLMs excel in many natural language processing tasks, they still face challenges, particularly the smaller models, in handling complex question-answering (QA) tasks \cite{selfask_bamboogle,iter_rag,yao2022react,xiong2024large,HZWVL+2024}. 

\begin{figure}
    \centering
    \includegraphics[width=0.5\textwidth]{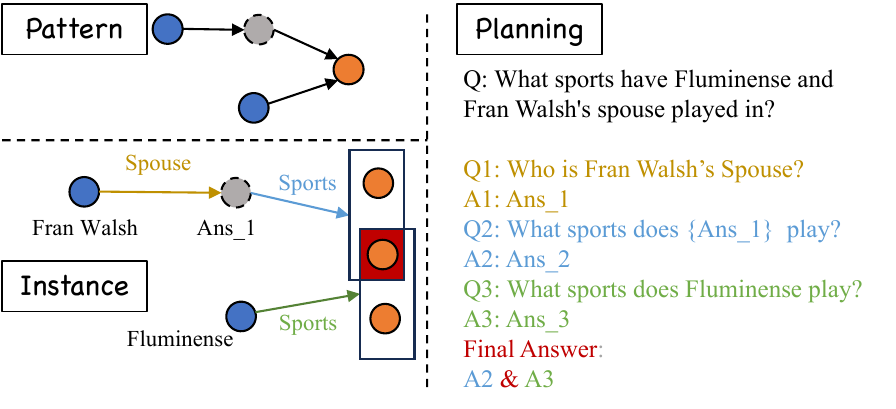}
    \caption{An example of a KG pattern, its grounded instance, and verbalized planning process.}
    \label{fig:kg pattern}
    % \vspace{-3mm}
\end{figure}

To improve the performance of LLMs on complex QA tasks, past research has tried various methods: (1) Employing carefully designed prompt strategies to guide the model in reasoning, such as Chain of Thought (CoT) \cite{zerocot,CoT} and Tree of Thought (ToT) \cite{yao2024tree} methods; (2) Utilizing retrieval techniques to obtain supplemental information from external knowledge source \cite{RAG,REALM}; (3) Combining prompt strategies with retrieval enhancements, as exemplified by methods like ReAct \cite{yao2022react} and Self-Ask \cite{selfask_bamboogle}. 
The third approach has garnered widespread research interest due to its integration of the advantages of the first two methods. 
The fundamental idea of this class of methods is to guide LLMs in breaking down a complex question into multiple simpler sub-questions and then use a retrieval-augmented generation (RAG)~\cite{HLVPP2023,HZWVL+2024} method to answer each sub-question, thereby deducing the answer to the original complex question. However, planning for complex questions is non-trivial, especially for smaller LLMs (with fewer than 10 billion parameters), which often require supervised fine-tuning \cite{aksitov2023rest,chen2023fireact,qin2023toolllm}.

This raises a widely concerning issue: how to obtain supervised data for learning the planning ability on complex questions. Manual annotation is time-consuming and labor-intensive, making it difficult to scale.
Most existing methods attempt to distill knowledge from teacher LLMs \cite{yao2022react,aksitov2023rest}, which places excessive trust in the teacher LLMs and, in reality, cannot guarantee the accuracy of the distilled knowledge. These challenges inspire us to explore new ways of obtaining supervised planning data.

Knowledge Graphs (KGs)~\cite{PVGW2017,PCEH+2017} usually store accurate knowledge in a structured way. 
We find that a KG pattern can be viewed as the abstract of a complex question, as shown in Figure \ref{fig:kg pattern}, which reveals the connection between question planning and patterns. 
% many questions can be abstract to the graph patterns (patterns in short), as shown in Figure \ref{fig:kg pattern}, which reveals the connection between question planning and patterns. 
% We find that many questions can be abstract to the graph patterns (patterns in short), as shown in Figure \ref{fig:kg pattern}, which reveals the connection between question planning and patterns. 
% The patterns \cite{GQE,ren2020beta} in KGs have brought us inspiration. 
% A complex question in natural language can be abstracted to an instance within a KG grounded from a pattern. 
% a pattern can be grounded as an instance within a KG, and this instance can then be verbalized into the planning for a complex question in natural language. 
This opens up the possibility of constructing training data to enhance the planning capabilities of LLMs using KGs.
% This process opens up the possibility of constructing supervision data for LLMs' planning. 
% Therefore, we attempt to utilize KGs to build planning data and use it to enhance the planning ability of LLMs. 
% As shown in Figure \ref{fig:kg pattern}, the patterns in KGs contain rich and correct reasoning paths, this opens up the possibility of constructing supervision data for LLMs' planning. 
Specifically, we start by grounding predefined patterns in an open-domain KG to extract numerous instances, which we then verbalize into complex questions and corresponding sub-questions in natural language. In this way, we effectively create a large number of accurate planning data for fine-tuning. 
% After fine-tuning LLMs with such planning data, 
% they are able to effectively generate plans for complex questions, 
Being fine-tuned with these planning data, 
LLMs' capability of generating plans for complex questions is enhanced, resulting in better final answers by parsing and executing these plans.
% allowing us to derive final answers through parsing and executing these plans.
We refer to this innovative framework as \textbf{L}earning to \textbf{P}lan from \textbf{K}nowledge \textbf{G}raphs (LPKG).

Additionally, we construct a  \textbf{C}omprehensive \textbf{L}ogical \textbf{QA} benchmark, CLQA-Wiki, from a subset of Wikidata \cite{Wikidata}  via grounding rich patterns as aforementioned. 
% we identify limitations in existing complex QA benchmarks \cite{yang2018hotpotqa,2wiki,selfask_bamboogle,trivedi2022musique}. They 
Existing complex QA benchmarks \cite{yang2018hotpotqa,2wiki,selfask_bamboogle,trivedi2022musique}
primarily focus on multi-hop and comparison-type questions and lack logical operations. Furthermore, most questions are labeled with only one answer, whereas in reality, they often have multiple correct answers.
% Conventional datasets are not built directly on KG patterns, they focus more on multi-hop and comparison questions. We hope to provide a more comprehensive evaluation of the logical reasoning ability of language models on different types of questions. 
% Thus, we construct a new \textbf{C}omprehensive \textbf{L}ogical \textbf{QA} benchmark, CLQA-Wiki, from a subset of Wikidata \cite{Wikidata}  via grounding rich patterns as aforementioned. 
The CLQA-Wiki benchmark evenly covers multi-hop, comparison, intersection, and union types of questions, 
which is more comprehensive and challenging for complex QA evaluation.
% providing a more comprehensive and challenging benchmark for complex QA evaluation.

Our contributions can be summarized as follows:
(1) We introduce a novel framework LPKG that enhances the planning ability of LLMs using data constructed from KG patterns;
(2) We develop a comprehensive and challenging evaluation benchmark, named CLQA-Wiki, to more effectively assess the performance of LLMs on complex QA tasks;
(3) Our proposed framework LPKG achieves better results than popular baselines on multiple conventional complex QA benchmarks, and we verify the effectiveness of the introduction of KG-sourced planning data.

\begin{figure*}
    \centering
    \includegraphics[width=\textwidth]{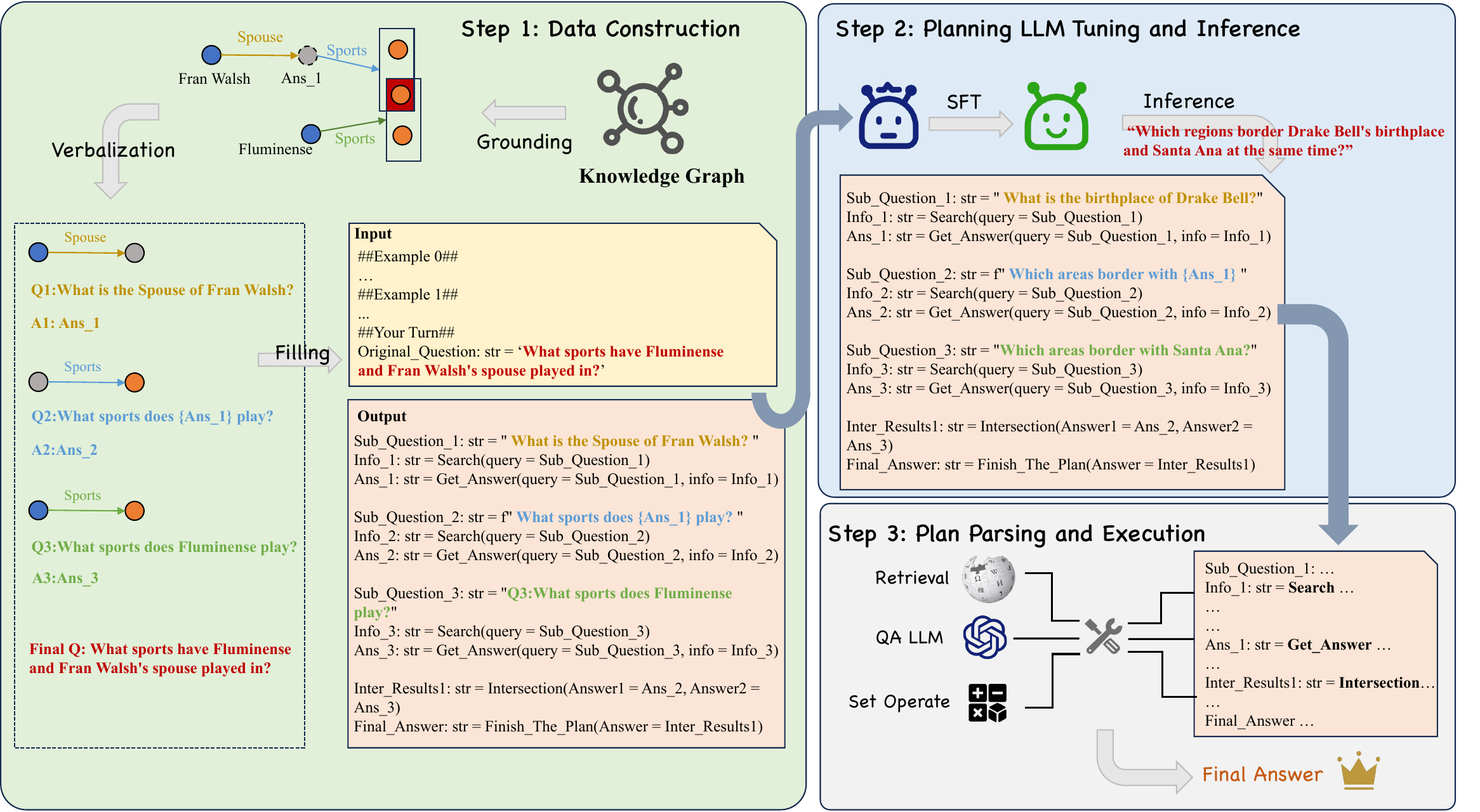}
    \caption{Overview of our Learning to Plan from Knowlege Graph ( LPKG) framework.}
    \label{fig:overview}
\end{figure*}
\section{Related Works}

% \subsection{Reasoning and Planning with LLMs}
\paragraph{Reasoning and Planning with LLMs}

In the context of LLMs, reasoning typically involves decomposing complex questions into sub-questions \cite{aug-llm-survey,RAP}. Prominent techniques include Chain-of-Thought (CoT) prompting \cite{CoT} which elicits rationales that lead to the final answers, and its extension, using self-consistency \cite{CoT-SC} or automated demonstration selection \cite{Auto-CoT}.
Other methods, such as ReAct \cite{yao2022react}, generate reasoning steps sequentially by integrating planning, with additional strategies like Tree of Thoughts (ToT) \cite{yao2024tree}, Reasoning via Planning (RAP) \cite{RAP}, and other methods \cite{DecompPrompt,least-to-most} facilitating complex question decomposition through varied planning approaches.
Unlike most methods that rely on in-context learning through prompt engineering, our approach generates planning data from KGs to fine-tune LLM, thereby enhancing their planning capabilities.

\paragraph{Retrieval-Augmented Generation}

Retrieval-Augmented Generation (RAG) can enhance LLMs by incorporating external data, allowing models to access up-to-date information and factual knowledge to mitigate hallucinations \cite{RAG-survey,REALM,RAG}. 
Each module in the RAG pipeline can be optimized, for instance, through retriever tuning \cite{replug,RA-DIT}, self-reflection during retrieval \cite{Self-RAG, CRAG}, or query refinement \cite{RQ-RAG}.
To address multi-hop questions, iterative RAG models \cite{iter_rag,ITRG,selfask_bamboogle} have been developed, which iteratively conduct retrieval-enhanced generation and generation-enhanced retrieval. However, the multiple RAG steps in existing methods are not optimized and rely heavily on in-context learning. Our approach uses planning data from KGs to facilitate more efficient RAG.

% Retrieval-Augmented Generation (RAG) is a paradigm that utilizes external information to enhance the generation ability of LLMs. Specifically, RAG retrieves relevant data from an external source for a question before sending it into a LLM.
% This paradigm has several advantages, like allow LLMs to access up-to-date information or fact knowledge for relieving hallucinations.
% RAG is usually treated as a long pipeline, and each step from query, retrieval, external information indexing to generation, can be optimized, like adding self reflection, 
% For handling multi-hop questions, ITER-RETGEN is proposed to iteratively conduct retrieval-enhanced generation and generation-enhanced retrieval. 

\paragraph{LLMs with KGs}

In the existing realm of LLMs, KGs are primarily utilized as sources of structured factual knowledge~\cite{PRKSC2023}. 
For example, Think-on-Graph \cite{ToG} extracts relevant triples from KGs to assist in QA.
Reasoning on Graph (RoG) \cite{RoG} generates relation-based plans and retrieves corresponding paths from these graphs. 
While aiding in KGQA tasks where answers are directly sourced from KGs, these graphs also support rationale generation. Chain-of-Knowledge (CoK) \cite{CoK} further leverages KGs along with other heterogeneous sources to generate faithful rationales.
Unlike previous studies, our approach constructs planning data for complex questions from KGs, recognizing that patterns within KGs inherently represent multi-step plans. This data is utilized to enhance the planning capabilities of LLMs.

\paragraph{Complex Logical Query in KGs}

Recent research on complex logic queries in KGs primarily focuses on first-order logical (FOL) queries that incorporate operations like conjunctions, disjunctions, negation, and existential quantifiers within incomplete KGs \cite{GQE,Query2Box,ren2020beta,CQD,FuzzQE,ENeSy,xiong2024teilp,WHHH+2024}.
These works define diverse patterns to assess the capability of logical operations in vector spaces, specifically targeting logical forms rather than natural language.
Nonetheless, their methodologies for pattern definition and extraction inspire our approach to deriving complex questions from KGs.

\section{Method}
\subsection{Overview}
% Figure 2 Work Flow
% Figure \ref{fig:overview} shows our overall framework. 
As shown in Figure \ref{fig:overview}, there are 3 steps in our \textbf{L}earning to \textbf{P}lan from \textbf{K}nowledge \textbf{G}raphs (LPKG) framework. (1) In the data construction step, we construct planning data from KGs. Specifically, we defined some basic KG patterns as shown in Figure \ref{fig:base_pattern}.
We ground patterns in an existing KG to extract instances. For each extracted instance, we sequentially verbalize the sub-queries within the instance into natural language sub-questions 
% (Please note that in this paper, we use ``query'' and ``question'' to refer to subgraph data and natural language data, respectively) 
according to their order in the instance, eventually assembling them into a complex question. Afterward, we build input and output templates for planning data, where complex questions are concatenated to the input prompt, and sub-questions are filled into the corresponding positions in the output text according to the type of patterns. 
(2) In the planning LLM tuning and inference step, we fine-tune LLMs based on such planning data to enable the LLMs to follow instructions to infer the plan for each question in the downstream test sets. 
(3) In the third step, such a plan will be parsed and executed, thereby obtaining the final answer to each question.
\subsection{Construction of Planning Data}
\begin{figure}
    \centering
    \includegraphics[width=0.48\textwidth]{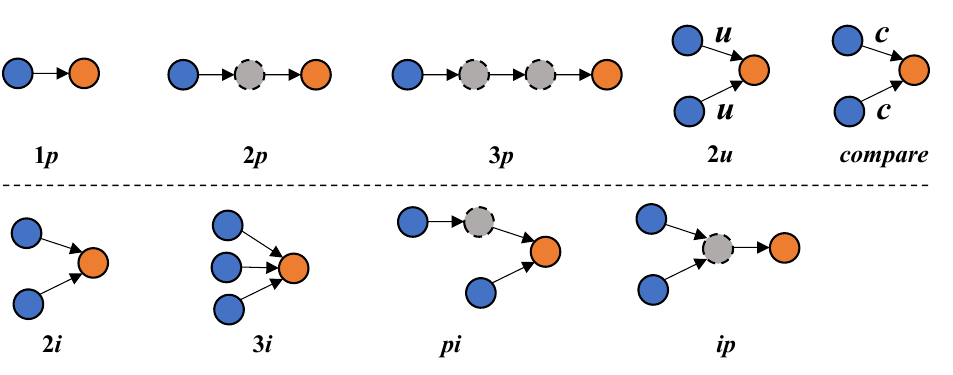}
    \caption{Basic KG patterns.}
    \label{fig:base_pattern}
\end{figure}
% \paragraph{Pattern Introduction.} 
\paragraph{Basic KG Patterns.} 
% Figure \ref{fig:base_pattern} shows predefined KG patterns in this work, which refer to previous work on complex logic query within KGs \cite{ren2020beta}. 
Inspired by previous work on complex logic queries within KGs \cite{ren2020beta}, we define the basic KG patterns as shown in Figure \ref{fig:base_pattern}. The set of KG patterns is denoted as $\mathcal{P}=\{1p,2p,3p,2i,3i,2u,ip,pi,compare\}$.
Specifically, $p,i,u$ respectively indicate projection, intersection, and union. 1$p$, 2$p$, and 3$p$ represent queries that span from one to three hops, 2$i$ and 3$i$ respectively represent the intersection of two sub-queries and three sub-queries, 2$u$ represents the union of two sub-queries, and $ip$ and $pi$ represent complex queries that combine two-hop with intersection logic. In addition, we also combine pairs of triples that have numeric tail entities and the same relations to construct comparison patterns, denoted as $compare$.

\paragraph{Grounding.} Given a KG, we first ground these patterns in it to extract instances:
\begin{equation}
    \mathcal{I}_{pat} = f_{pat}(\mathcal{KG}), pat \in \mathcal{P}
\end{equation}
where $\mathcal{I}_{pat}$ are the instances grounded by knowledge graph $\mathcal{KG}$ of pattern $pat$, ${f_{pat}}$ is the corresponding extraction function. For example, an instance of the $2p$ pattern can be ``(Inkheart, (cast member, educated at))''.  
% and $\mathcal{P}$ is a pattern set contains all the patterns we need. 
To best meet the needs of open-domain QA, we use Wikidata15k \cite{wikidata15k}, a subset of the open-domain KG Wikidata, as $\mathcal{KG}$.
% To cover as wide a variety of real-world questions as possible, 9 types of patterns in $\mathcal{P}$ are identified, i.e., $\mathcal{P}=\{1p,2p,3p,2i,3i,2u,ip,pi,compare\}$ . 

\paragraph{Verbalization.} Subsequently, based on the grounded instances, we need to verbalize them bottom-up into sub-questions and assemble them into complex questions. There are several methods for this step, such as a templates-based method, manual annotation, or utilizing an LLM. Since the template-based approach often lacks fluency in language expression, and the manual method is time-consuming and labor-intensive, we opt for an LLM-based method. Specifically, we write a small number of verbalization examples for each pattern type. These examples are used as demonstrations $De_{1}$ to fill in the prompt. Finally, we concatenate a grounded instance $i \in \mathcal{I}_{pat}$ to the prompt, asking an LLM to verbalize it to a natural language question:
\begin{equation}
    \{\{Q_{s_n}\}_{n=1}^k,Q_{c}\} = llm(concat(De_{1},i))
\end{equation}
where $\{Q_{s_n}\}_{n=1}^k \text{ and } Q_{c}$ represent the resulting sub-questions and complex question respectively, $concat$ is string level concatenation. We use GPT-4 as $llm$ here. It is important to note that here the $llm$'s role is merely to transform the data format; the sub-questions and complex question still originate from the structure of the KG itself, without introducing any knowledge from the $llm$ in the task of question planning. The prompt we use can be found in Appendix \ref{verb prompt}.

\paragraph{Filling.} We then extract sub-questions and complex questions from the output of the $llm$. Subsequently, we built a set of planning templates $\mathcal{T}_{pat}$ for the planning process of questions corresponding to each pattern. The $\{Q_{s_n}\}_{n=1}^k$ obtained in the previous step will be filled into fixed positions in $\mathcal{T}_{pat}$ corresponding to their pattern type, thereby obtaining the output for training. The $Q_{c}$ obtained in the previous step is concatenated to the end of a fixed instruction $Ins$ and some planning demonstrations $De_{2}$ (also constructed from KGs), thus obtaining the input for training data:
\begin{equation}\label{input}
    x = concat(Ins, De_{2}, Q_{c})
\end{equation}
\begin{equation}
    y = \mathcal{T}_{pat}.fill(\{Q_{s}\}_{n=1}^k), pat \in \mathcal{P}
\end{equation}
where $.fill$ is a filling function of templates $\mathcal{T}_{pat}$. Inspired by \cite{aksitov2023rest}, we use a code-formatted input $x \text{ and output } y$ here (shown in ``Input'' and ``Output'' in Figure \ref{fig:overview}) to facilitate formatting and subsequent parsing and execution of the output plan (more details in Appendix \ref{planning prompt}).
In the end, we obtain 9000 training data entries $\mathcal{D}_{train} = \{x_n,y_n\}_{n=1}^{9000}$, with 1000 entries for each pattern. We randomly select 100 items from the training sets for manual verification, with an accuracy rate of over 95\%.

\subsection{Fine-tuning and Inference of Planning LLMs}

We use the obtained training data $\mathcal{D}_{train}$ to fine-tune the planning LLMs ${\mathcal{M}_p}$ directly with the standard next token training objective:
\begin{equation}
    \mathop{max}\limits_{\mathcal{M}_p} \mathbb{E}_{(x,y)\in\mathcal{D}_{train}} \text{ Log }p_{\mathcal{M}_p}(y|x)
\end{equation}

The fine-tuned planning LLM $\mathcal{M}_p$ can be used to infer the plan $P$ for each question $Q_{test}$ in the downstream test set:
\begin{equation}\label{testprompt}
    P = \mathcal{M}_p(concat(Ins,De_2,Q_{test}))
\end{equation}
where $Ins$ and $De_2$ are the same as the contents in the Equation (\ref{input}). It should be noted that in the multi-hop questions, the specific sub-questions in the second and third hops need to be constructed based on the answers to the previous hop's sub-questions. Since our $P$ outputs all processes at once, the $\mathcal{M}_p$ cannot know the answers to the previous hop's sub-questions when outputting the plans. Therefore, we will use a placeholder to replace the answer to the previous hop sub-questions, allowing the planning to proceed smoothly (as shown in Table \ref{tab:2p}, \ref{tab:3p}, \ref{tab:ip}, \ref{tab:pi} in Appendix \ref{verb prompt}). These placeholders will then be filled in during the subsequent parsing and execution process.

\subsection{Plan Parsing and Execution}
The obtained plan $P$ needs to be parsed and executed to obtain the final answer of the $Q_{test}$. Due to our adoption of code-formatted input and output for fine-tuning the $\mathcal{M}_p$, the $P$ here is also highly formatted code, which facilitates our parsing of each step of the plan and executing them. In particular:

$\bullet$ When a step includes a ``Search'' function, we will call an external retrieval tool.

$\bullet$ When a step includes a ``Get Answer'' function, we'll invoke an external QA LLM $\mathcal{M}_{QA}$ to get answers for a sub-question based on the retrieved information. The possible placeholders in sub-questions will be filled with previous answers. We ask QA LLM to organize answers in the form of a list (prompt is shown in Table \ref{tab:qa_llm} in Appendix \ref{qa prompt}).

$\bullet$ When ``Intersection'' or ``Union'' appears in the step, we will run actual intersection or union functions. This can be easily completed due to list format answers in the previous step.

It is important to note that the planning LLM $\mathcal{M}_p$ and the QA LLM $\mathcal{M}_{QA}$ are completely decoupled in our framework. Here we can use any LLM off-the-shelf to handle the task of QA. Ultimately, we can obtain the answer to $Q_{test}$.
\section{New BenchMark: CLQA-Wiki}
\begin{table}[t]
    \centering
    % \vspace{-3mm}
    % \small
    \resizebox{\columnwidth}{!}{
    \begin{tabular}{c|c|c|c}
    \toprule
    Type & Count & Type & Count \\
    \hline
      2p question & 200 & 3p question & 200 \\
      2i question & 200 & 3i question & 200\\
      ip question & 50 & pi question & 50 \\
      2u question & 200 & compare question & 100\\
    \bottomrule
    \end{tabular}
    }
    \caption{Distribution of CLQA-Wiki.}
    % \vspace{-3mm}
    \label{tab:distribute}
\end{table}
The conventional complex QA datasets include HotPotQA \cite{yang2018hotpotqa}, 2WikiMultihopQA \cite{2wiki}, MuSiQue \cite{trivedi2022musique}, and Bamboogle \cite{selfask_bamboogle}. Despite their widespread use in evaluating the QA performance of language models, we identify some problems with these datasets:

(1) All these datasets are primarily focused on multi-hop and comparison-type questions. The types of questions are not balanced and comprehensive enough, and less attention is paid to questions involving intersection and union logic,  which are also very common in reality. 
% A typical example of an intersection question is, ``Which country borders with Russia and China at the same time?''

(2) Except for MuSiQue, the questions on the rest of the other three datasets only have one answer, whereas many questions in reality often have multiple answers. For example, the answer to an intersection question ``Which country borders with Russia and China at the same time?'' is a set [Mongolia, Kazakhstan, North Korea].

In light of this, we aim to construct a new testing benchmark that embodies more comprehensive logic and allows for an unrestricted number of answers to more thoroughly evaluate the performance of language models on various logical questions. Considering the detailed pattern structures and unrestricted number of answer entities in KGs, we construct a test set based on Wikidata15k.

Similar to the method used to construct the planning data, we extract instances from Wikidata15k (which do not appear in the training data) and use GPT-4 to do verbalization. Moreover, for each instance, we can obtain all the answer entities from Wikidata15k, which we then designate as the answers to the questions. After manual quality checks, we obtain a test set called CLQA-Wiki, which contains 1,200 pieces of data featuring a variety of \textbf{C}omprehensive \textbf{L}ogical \textbf{QA} pairs. The question types and their distribution are listed in Table \ref{tab:distribute}. It is worth noting that we have constructed 9 types of testing questions until now, and for newly defined patterns, we can also quickly construct corresponding questions using the above method, showing the better scalability of our dataset.

\section{Experiment}
We aim to answer the following research questions in our experiments:

$\bullet$ \textbf{RQ1}: Can LPKG outperform baseline methods on conventional complex QA datasets?

$\bullet$ \textbf{RQ2}: Can planning data derived from KGs help improve the planning ability of the LLMs?

$\bullet$ \textbf{RQ3}: Can planning data derived from KGs be more helpful in improving the LLMs' planning ability compared to normal distillation methods?

$\bullet$ \textbf{RQ4}: Can LPKG outperform baseline methods on the new benchmark CLQA-Wiki? 
\begin{table*}[t]
    \centering
    % \vspace{-3mm}
    % \small
    \begin{tabular}{l|c|c|c|c|c|c }
    \toprule
    &Planning&RAG&HotPotQA & 2WikiMQA & Bamboogle & MuSiQue \\
    \hline
      {Direct} & \XSolidBrush & \XSolidBrush & {0.268} &  {0.284} &  {0.128} &  {0.090} \\
      {CoT} & \CheckmarkBold & \XSolidBrush & {0.288} &  {0.286} &  {0.280} &  {0.090}\\
      {Direct RAG} & \XSolidBrush & \CheckmarkBold & {0.292} &  {0.230} &  {0.080} &  {0.088}\\
      % \hline
      % \multicolumn{5}{l}{\textit{Combine Planning and Retrieval}}\\
      \hline
      ReAct & \CheckmarkBold & \CheckmarkBold & {0.211} &  {0.216} &  {0.168} &  {0.060} \\
      Self-Ask & \CheckmarkBold & \CheckmarkBold &{0.176} &  {0.194} &  {0.136} &  {0.116} \\
      % GPT-3.5 $\mathcal{M}_p$ (>175B) + GPT-3.5 RAG & \underline{0.352} &  {0.344} &  \underline{0.272} &  {0.254} \\
      % \hline
      % \multicolumn{5}{l}{\textit{Ours}}\\
      \hline
      {ICLPKG(GPT-3.5)} & \CheckmarkBold & \CheckmarkBold & \underline{0.352} &  {0.344} &  \underline{0.296} &  {0.254} \\
      {LPKG(CodeQwen)}& \CheckmarkBold & \CheckmarkBold & {0.338} &  \underline{0.356} &  {0.280} &  \underline{0.266} \\
      {LPKG(Llama3)}& \CheckmarkBold & \CheckmarkBold & \textbf{0.376} &  \textbf{0.372} &  \textbf{0.304} &  \textbf{0.296} \\
    \bottomrule
    \end{tabular}
    \caption{Exact match results on conventional complex QA datasets. The best results are in bold, and the second best is underlined. All baseline methods are conducted on GPT-3.5. LPKG(CodeQwen), and LPKG(Llama3) respectively represent using our framework with fine-tuned CodeQwen1.5-7B-Chat and fine-tuned Llama3-8B-Instruct (fine-tuning is conducted on KG-sourced planning data).}
    % \vspace{-3mm}
    \label{tab:main_results}
\end{table*}
\begin{table*}[t]
    \centering
    % \vspace{-3mm}
    % \small
    \begin{tabular}{c|c|c|c|c }
    \toprule
     & HotPotQA & 2WikiMQA & Bamboogle & MuSiQue \\
    \hline
     LPKG(CodeQwen) & \textbf{0.338} &  \textbf{0.356} &  \textbf{0.280} &  \textbf{0.266} \\
      ICLPKG(CodeQwen)  & {0.110} &  {0.286} &  {0.192} &  {0.176} \\
    \hline
      LPKG(Llama3)  & \textbf{0.376} &  \textbf{0.372} &  \textbf{0.304} &  \textbf{0.296} \\
      ICLPKG(Llama3)  & {0.369} &  {0.353} &  {0.280} &  {0.290} \\
    \bottomrule
    \end{tabular}
    \caption{Ablation study on the  KG-sourced planning data. ICLPKG(CodeQwen) and ICLPKG(Llama3) represent using the raw CodeQwen1.5-7B-Chat and Llama3-8B-Instruct to conduct planning, respectively.}
    % \vspace{-3mm}
    \label{tab:ab1}
\end{table*}

\subsection{Experimental Settings}

\paragraph{Datasets}
We first conduct experiments on four conventional complex QA datasets: HotPotQA \cite{yang2018hotpotqa}, 2WikiMultiHopQA(2WikiMQA) \cite{2wiki}, MuSiQue \cite{trivedi2022musique}, and Bamboogle \cite{selfask_bamboogle}. Among them, HotPotQA, 2WikiMQA, and MuSiQue contain completed train sets, development sets, and test sets, while Bamboogle is a small dataset that only contains 125 test data. Similar to the previous method \cite{iter_rag,aksitov2023rest}, we respectively extract the first 500 entries from the development set of HotPotQA, 2WikiMQA. For MuSiQue, we follow \citet{selfask_bamboogle} to use only 2-hop questions in the development set. And for Bamboogle, we use all of its data as test data. Finally, we conduct testing on our benchmark CLQA-Wiki.

\paragraph{Baselines}
We compare our framework to various baselines:
$\bullet$ \textbf{Direct}: Directly input the original question into LLM.
$\bullet$ \textbf{CoT}: Follow \citet{zerocot}, we instruct LLM firstly ``Think step by step'' and then give the final answers.
$\bullet$ \textbf{Direct RAG}: The prompt sent to LLM contains the original question and retrieved information related to the original question.
$\bullet$ \textbf{ReAct} \cite{yao2022react}: Answering questions through iterative planning, action, and observation. The action here is the retrieval tool and observation is the retrieved information. The planning and QA are conducted on a single LLM.
$\bullet$ \textbf{Self-Ask} \cite{selfask_bamboogle}: Similar to ReAct, it first instructs LLM to judge whether sub-questions are needed. If so, it will request LLM to generate the sub-questions, then conduct external retrieval based on the sub-questions, and allow LLM to provide answers based on the retrieved information. $\bullet$ \textbf{ICLPKG} A variant of LPKG framework. Planning LLMs are not fine-tuned, while just using \textbf{I}n-\textbf{C}ontext \textbf{L}earning to do \textbf{P}lanning with some \textbf{KG}-sourced planning demonstrations.

\paragraph{Evaluation Metrics}
Exact Match (EM) is set as an evaluation metric in HotPotQA, 2WikiMQA, Bamboogle, and MuSiQue. While in CLQA-Wiki, we use Recall and Precision.

\paragraph{Implementation Details} All baselines are conducted with \texttt{gpt-3.5-turbo-1106}\footnote{https://platform.openai.com/docs/models/gpt-3-5-turbo} (GPT-3.5). The prompts of ``Direct'', ``CoT'', and ``Direct RAG'' are written by ourselves. The ReAct and Self-Ask are replicated based on their source code with the GPT-3.5 API. To facilitate assessment, we will ask the model to only output concise answer phrases.

In our framework: (1) For pattern grounding, we use Wikidata15k as $\mathcal{KG}$, which contains about 15k entities and 263 relations. The extraction tool in grounding is modified from existing works \cite{ren2020beta}.
(2) For the planning LLM $\mathcal{M}_p$, we choose CodeQwen1.5-7B-Chat and Llama3-8B-Instruct, one excels at coding while the other excels at common sense reasoning. We fine-tune them with Lora tuning, running on 4x80G A100 GPUs for about 3 hours. The fine-tuning is conducted for 2 epochs, with a learning rate of 5$e$-5 and a cosine learning rate scheduler.
(3) For retrieval, following previous works \cite{iter_rag,Self-RAG}, we employ Wikipedia as the corpus for document retrieval and use the off-the-shelf Contriever-MS as the retriever. We select the Top 5 documents as the retrieved information.
(4) For QA LLM, since we only care about the ability of the planning LLMs, in order to eliminate the impact of differences in the ability of QA LLMs, we use GPT-3.5 to align with baselines. 

\subsection{Results on Conventional Complex QA}

\paragraph{Main Results (RQ1,RQ2)}
Table \ref{tab:main_results} shows results on conventional complex QA datasets. Since the QA LLM remains unchanged in our framework, we use ``LPKG(Llama3)'' and ``LPKG(CodeQwen)'' to represent LPKG frameworks using different planning LLMs, respectively. They are fine-tuned on Llama3-8B-Instruct and CodeQwen1.5-7B-Chat with KG-sourced planning data. It can be found that our framework outperforms the baseline methods on the majority of datasets. Particularly, compared to ReAct and Self-Ask, our approach shows significant improvement. It is worth noting that both ReAct and Self-Ask iterative planning and RAG in their workflows, whereas our approach decouples planning and RAG into two separate models. This allows each model to focus more intensively on its individual task. Moreover, we specifically enhance the planning part by fine-tuning $\mathcal{M}_p$ with planning data sourced from KG. These two changes have brought significant improvements to the overall accuracy.

At the same time, we also attempt to replace the fine-tuned $\mathcal{M}_p$ with GPT-3.5 while keeping other parts unchanged, denoted as ``ICLPKG(GPT-3.5)'' in Table \ref{tab:main_results}. Results show that even though fine-tuned $\mathcal{M}_p$ CodeQwen (7B) and Llama3 (8B) have significantly fewer parameters than GPT-3.5 (more than 175B), they can maintain or even surpass GPT-3.5 in terms of planning ability. Next, we replace the $\mathcal{M}_p$ fine-tuned on KG-sourced data with their raw models, and the experimental results are shown in Table \ref{tab:ab1}. It can be observed that after fine-tuning with planning data derived from the KG, both CodeQwen1.5-7B-Chat and Llama3-8B-Instruct show significant improvements in planning ability. In particular, CodeQwen1.5-7B-Chat, which is significantly inferior to GPT-3.5 across all datasets in planning ability before fine-tuning, exhibits a notable enhancement after fine-tuning on KG-based planning data, especially achieving better results than GPT-3.5 on 2WikiMQA and MuSiQue. All these experimental phenomena fully demonstrate the efficacy of using KG-sourced planning data in improving the planning ability of the LLMs.

\paragraph{Compare to Normal Distillation (RQ3)}
\begin{table}[t]
    \centering
    % \vspace{-3mm}
    % \small
    \begin{tabular}{l|c}
    \toprule
    & Bamboogle \\
    \hline
      LPKG(CodeQwen) & \textbf{0.280}\\
      DLPKG(CodeQwen)  & \underline{0.216}\\
      ICLPKG(CodeQwen) & {0.192}\\
    % \hline
    \bottomrule
    \end{tabular}
    \caption{Comparison with normal distillation methods. QA LLM is GPT-3.5.}
    % \vspace{-3mm}
    \label{tab:distill}
\end{table}

To further validate the effectiveness of using planning data constructed from KG, we compare it with the normal distillation method. Specifically, we extracted 3000 questions each from the training sets of HotPotQA, 2WikiMQA, and MuSiQue (9000 questions in total). Using the same input prompt with Equation (\ref{testprompt}), we obtain the planning process of these questions by invoking GPT-3.5. These planning data are then used to fine-tune CodeQwen1.5-7B-Chat which has relatively weaker planning capabilities, resulting in DLPKG(CodeQwen).

To ensure the fairness of the comparison, we conduct testing on the unseen dataset Bamboogle, and the experimental results are shown in Table \ref{tab:distill}. The results demonstrate that, under the same amount of training data and without using in-domain questions for fine-tuning, using planning data constructed from KG yields better performance than using planning data distilled from GPT-3.5. We believe this observation is inspiring and can be attributed to the richer reasoning types in the KG patterns, as well as the highly accurate reasoning paths in well-constructed KG.

\paragraph{Error Analysis}
\begin{table}[t]
    \centering
    % \vspace{-3mm}
    % \small
    \resizebox{\columnwidth}{!}{
    \begin{tabular}{c|c|c}
    \toprule
    Planning Error & Retrieval Error & QA LLM  Error\\
    \hline
    13 & 17 & 10\\
    \bottomrule
    \end{tabular}
    }
    \caption{Error analysis of LPKG(Llama3).}
    % \vspace{-3mm}
    \label{tab:error_analysis}
\end{table}
To gain a deeper understanding of the model's performance, we conduct an error analysis of LPKG. Specifically, we extract 40 incorrect samples (10 per dataset) of LPKG(Llama3) and manually categorize the error cases into three types: planning error, retrieval error, and QA LLM error. As shown in Table \ref{tab:error_analysis}, the performance of the retrieval model has the greatest impact. Among the 13 samples with planning errors, 10 of them are due to incorrect judgment of the type of questions, and 3 are due to incorrect expression of sub-questions.
Future exploration directions can be based on improving the performance of the retriever model and enhancing the planning LLM's ability to identify question types. 
\subsection{Results on CLQA-Wiki (RQ4)}

\paragraph{Main Results}
\begin{table}[t]
    \centering
    % \vspace{-3mm}
    % \small
    \resizebox{\columnwidth}{!}{
    \begin{tabular}{l|l|l }
    \toprule
    & \multicolumn{2}{c}{CLQA-Wiki}\\
    \cline{2-3}
    & Precision & Recall \\
    \hline
      \small{CoT} &  {0.0605\tiny{\textcolor{red}{(+80.6\%)}}} &  {0.0641\tiny{\textcolor{red}{(+103.4\%)}}} \\
      \hline
      \small{Direct RAG} &  {0.0814\tiny{\textcolor{red}{(+34.2\%)}}} &  {0.0789\tiny{\textcolor{red}{(+65.3\%)}}} \\
      \hline
      \small{ReAct} & {0.0264\tiny{\textcolor{red}{(+314.0\%)}}} &  {0.0270\tiny{\textcolor{red}{(+382.9\%)}}} \\
      \hline
      \small{Self-Ask} & {0.0385\tiny{\textcolor{red}{(+183.8\%)}}} &  {0.0423\tiny{\textcolor{red}{(+208.2\%)}}} \\
      \hline
      % \multicolumn{3}{l}{Ours ($\mathcal{M}_p$ + GPT3.5 RAG)} \\
      % \hline
      % \tabincell{l}{GPT-3.5 $\mathcal{M}_p$\\+\\GPT-3.5 RAG}
      \small{ICLPKG(GPT-3.5)}
      & {0.0907\tiny{\textcolor{red}{(+20.5\%)}}} &  {0.1014\tiny{\textcolor{red}{(+28.6\%)}}} \\
      \hline
      % \multicolumn{3}{l}{\textit{Ours}}\\
      % \hline
      % \tabincell{l}{Llama3-KGP $\mathcal{M}_p$\\+\\GPT-3.5 RAG} 
      \small{LPKG(Llama3)}
      & \textbf{0.1112} &  \textbf{0.1344} \\
    \bottomrule
    \end{tabular}
    }
    \caption{Precision and Recall result on CLQA-Wiki.}
    % \vspace{-3mm}
    \label{tab:clqa_results}
\end{table}
\begin{figure*}
    \centering
    \includegraphics[width=\textwidth]{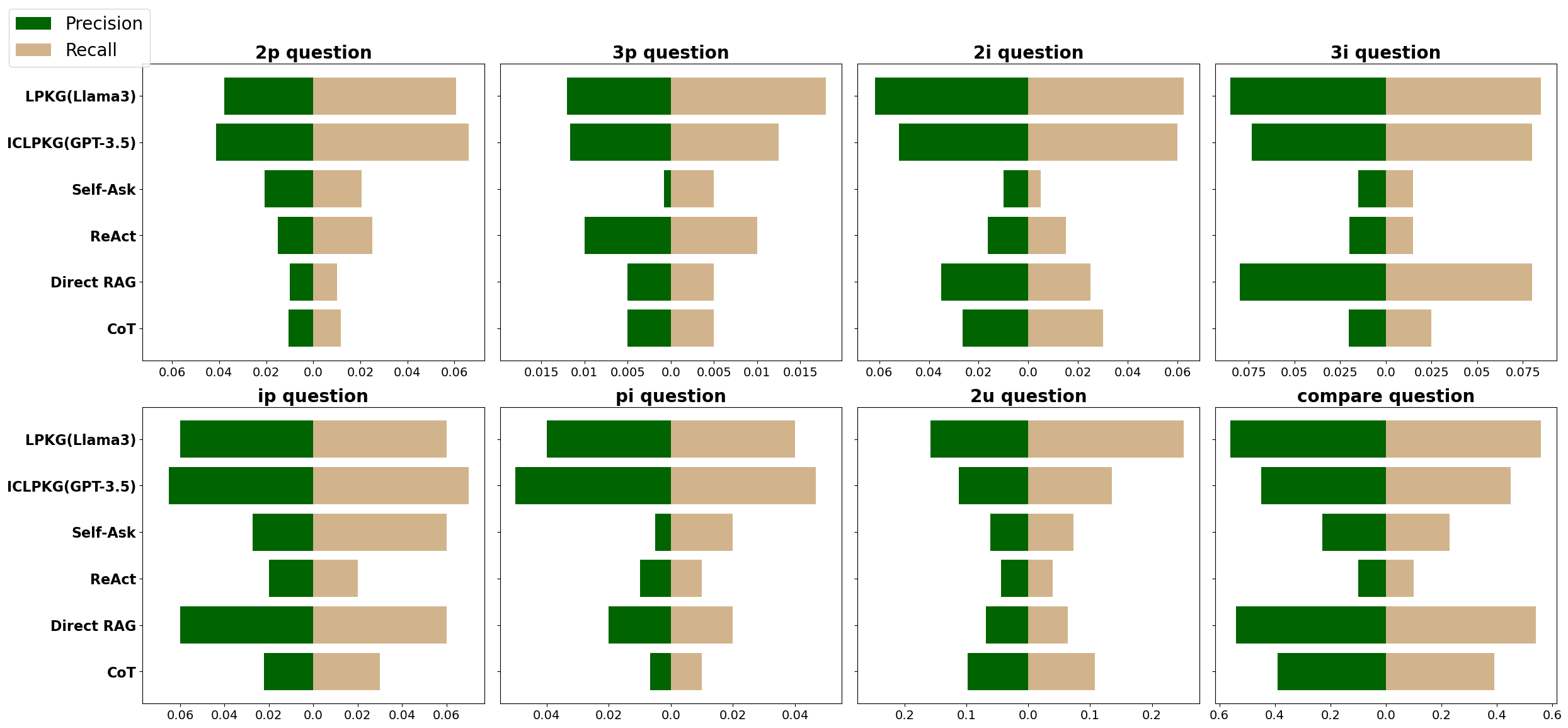}
    \caption{Fine-grained evaluation based on question types.}
    \label{fig:fine_grained}
\end{figure*}

We then conduct testing based on the CLQA-Wiki benchmark. Given that answers in this benchmark may have multiple candidates, we adjust the instructions for QA LLMs to require them to output all potential answers in a specified list format. This adjustment is made to facilitate the extraction and evaluation of the responses. Since Llama3-8B-Instruct is more powerful than CodeQwen1.5-7B-Chat as shown in Table \ref{tab:main_results}, we only conduct LPKG with Llama3 here.
Experimental results are presented in Table \ref{tab:clqa_results}. It can be seen that CLQA-Wiki is a very challenging dataset, but LPKG(Llama3) still outperforms the baseline methods. At the same time, compared to ICLPKG(GPT-3.5), LPKG(Llama3) has an average improvement of over 20\%, highlighting the importance of using KG-sourced planning data. 

In addition, we conduct more fine-grained experiments based on the type of questions, and the experimental results are shown in Figure \ref{fig:fine_grained}. We found that LPKG(Llama3) performs more prominently on some complex questions, such as the 3$p$, 2$i$, and 2$u$ questions, demonstrating the advantages of our framework in dealing with complex logic questions. At the same time, we also found that direct retrieval performs well on some types of questions, such as 3$i$ and compare questions. This may be due to the fact that in the process of verbalizing these questions, the assembly of sub-questions into complex questions is relatively straightforward, allowing answers to each sub-questions to be obtained directly through the retrieval of complex questions or the knowledge of the LLM itself.

\paragraph{Case Study}
To more intuitively demonstrate the effectiveness of KG-source planning data, we conduct a case study on CLQA-Wiki, detailed in Figure \ref{fig:case_study} in Appendix \ref{sec:case}. When planning a 2$i$ question ``What sport is associated with John Madden and Ben Johnson?'', GPT-3.5 generates some meaningless sub-questions and incorrectly defines the question type, which will definitely lead to incorrect answers. But LPKG(Llama3) could identify it as a 2$i$ question and provide the correct sub-questions and planning steps,  thereby helping to obtain the correct answer during final parsing and execution.

\section{Conclusion}
In this paper, we try to enhance the planning ability in retrieval-augmented LLMs using KGs. Specifically, we design a framework for Learning to Plan from KG (LPKG). The proposed LPKG framework first utilizes the rich patterns in the KGs to construct planning data, then fine-tune planning LLMs based on such data to enable them to conduct planning on downstream datasets, and ultimately get the final answer through parsing and execution. The experimental results reveal the excellent performance of the LPKG framework and also demonstrate the effectiveness of using KG-sourced data to enhance LLMs' planning ability. Finally, we construct CLQA-Wiki, providing a more challenging complex QA benchmark for the community.

% \clearpage
\section*{Limitation}
In our view, the limitations of our work at the current stage mainly stem from two aspects:

(1) During the fine-tuning phase of planning LLMs, we simply mixed various types of questions together uniformly for training. We have not yet explored the impact of question type distribution on the experimental results, which could be the focus of future work.

(2) At present, the datasets we test have explicit types of questions (multi-hop, comparison, and union/intersection), but in reality, some question types may be implicit or even not be included in the types we define. The future direction of our work can be to study planning methods for these types of unclear questions.

\section*{Acknowledgement}
This work is founded by National Natural Science Foundation of China (NSFC62306276/NSFCU23B2055/NSFCU19B2027), Zhejiang Provincial Natural Science Foundation of China (No. LQ23F020017), Yongjiang Talent Introduction Programme (2022A-238-G), Chang Jiang Scholars Program (J2019032), Fundamental Research Funds for the Central Universities (226-2023-00138).
\bibliography{custom}

\clearpage
\appendix

\section{Case Study}
\label{sec:case}

\begin{figure*}
    \centering
    \includegraphics[width=\textwidth]{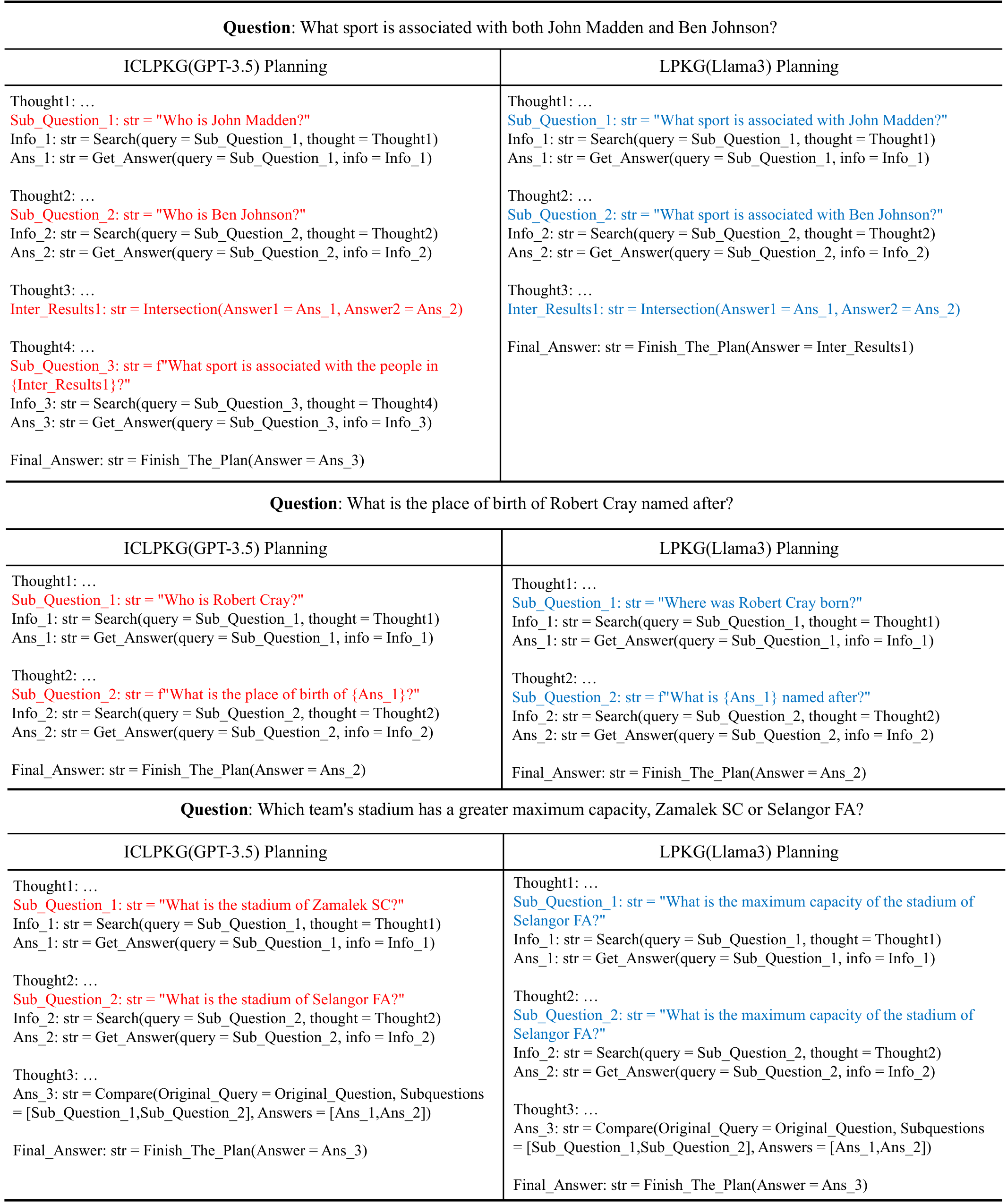}
    \caption{Case study on CLQA-Wiki.}
    \label{fig:case_study}
\end{figure*}

Cases are shown in Figure \ref{fig:case_study} where we mark the bad parts of the ICLPKG (GPT-3.5) planning in red, and the corresponding parts of LPKG (Llama3) in blue. In order to highlight the core difference, the ``Thought'' in planning is omitted in the cases.
\section{Prevention of Data Leakage}
In the experiment, we used data from knowledge graph sources as training data, which may raise concerns about data leakage, specifically the overlap between the training data and the four multi-hop test sets(HotPotQA, 2WikiMQA, Bamboogle, MusiQue). We calculate the semantic similarity between the training and testing questions. We use the BGE-M3 model to embed the training and testing questions into high-dimensional vectors and calculate their cosine similarity. We found that the similarity between the testing and training questions does not exceed 0.8, and is generally below 0.6. This indicates that there is almost no high degree of overlap between the training data and the testing data.

In addition, when conducting experiments on our own test set CLQA-Wiki, we excluded questions similar to those in CLQA-Wiki from the training data (with similarity scores above 0.9), ensuring no high overlap between the training and testing data.

\section{Prompt Content}
\subsection{Prompt of Verbalization}
\label{verb prompt}
Table \ref{tab:1p},\ref{tab:2p},\ref{tab:3p},\ref{tab:2i},\ref{tab:3i},\ref{tab:ip},\ref{tab:pi},\ref{tab:2u},\ref{tab:compare} shows the different prompts of verbalizing pattern instances to their natural language questions. The specific instance that needs to be verbalized will be added to the end of the prompt.

\subsection{Prompt of Planning LLM}
\label{planning prompt}
The code-formatted input prompt for planning LLM is as follows. Due to space limitations, only some demonstrations are displayed in the prompt. In fact, we will include planning demonstrations for different types of questions in the prompt:

\begin{lstlisting}[basicstyle = \scriptsize\ttfamily, breaklines = true]

###Complete the Code Below###

from package1 import SerpAPIWrapper
from package2 import QA_LLM
search = SerpAPIWrapper()

def Search(query:str,thought:str):
    """Search relevant information about query based on external Search Engine.
    Attributes:
		query: The question you want to search.
		thought: The reason why this query is need. 
    """
    if thought is not None:
        return search.run(query)
    else:
        return ("Please give your thought!")

def Get_Answer(query:str,info:str):
    """Get the answer of the query based on the information.
    Attributes:
    query: The question you want to search.
    info: The information relevant to the query.
    """
    ### Use the QA_LLM model to get the answer.
    return QA_LLM(query,info)

def Compare(Original_Query:str,Subquestions:list,Answers:list):
    """Compare the answer of the sub-questions and return the final answer of original query.
    Attributes:
    Original_Query: The original question.
    Subquestions: The list of sub-questions.
    Answers: The list of answers of the sub-questions.
    """
    query = Original_Query
    info = str()
    for i in range(len(Subquestions)):
        info += Subquestions[i] + ' : ' + Answers[i] + '\n'
    return QA_LLM(query,info)

def Intersection(Answer1:str,Answer2:str):
    """Find the intersection of two answer sets.
    Attributes:
    Answer1: The first answer set.
    Answer2: The second answer set.
    """
    List1 = Answer1.split(',')
    List2 = Answer2.split(',')
    return str(set(List1) & set(List2))

def Union(Answer1:str,Answer2:str):
    """Find the union of two answer sets.
    Attributes:
    Answer1: The first answer set.
    Answer2: The second answer set.
    """
    List1 = Answer1.split(',')
    List2 = Answer2.split(',')
    return str(set(List1) | set(List2))

def Finish_The_Plan(Answer:str):
    """Call this function to finish the plan and return the final answer.
    Attributes:
    Answer: The final answer of the original question.
    """
    return Answer

###################
# Example 0:
###################

Original_Question: str = "What is the ethnic group of Booker T. Jones?"
### Question Type: One Projection
### Decompose the original question into sub-questions.

Thought1: str = "An atomic question, no need to decompose. Search directly."
Sub_Question_1: str = "What is the ethnic group of Booker T. Jones?"
Info_1: str = Search(query = Sub_Question_1, thought = Thought1)
Ans_1: str = Get_Answer(query = Sub_Question_1, info = Info_1)

Final_Answer: str = Finish_The_Plan(Answer = Ans_1)

###################
# Example 1:
###################

Original_Question: str = "Who succeeded the first President of Namibia?"
### Question Type: Two Projection
### Decompose the original question into sub-questions.

Thought1: str = "If I want to know who succeeded the first President of Namibia, I need to first know who is the first President of Namibia."
Sub_Question_1: str = "Who is the first President of Namibia?"
Info_1: str = Search(query = Sub_Question_1, thought = Thought1)
Ans_1: str = Get_Answer(query = Sub_Question_1, info = Info_1)

Thought2: str = "After knowing who is the first President of Namibia, I need to know who succeeded him."
Sub_Question_2: str = f"Who succeeded {Ans_1}?"
Info_2: str = Search(query = Sub_Question_2, thought = Thought2)
Ans_2: str = Get_Answer(query = Sub_Question_2, info = Info_2)

Final_Answer: str = Finish_The_Plan(Answer = Ans_2)

......(More Examples are omitted here)
###################
# Your turn! Just complete the code below and do not return other things.
###################

Original_Question: str = 

\end{lstlisting}

\subsection{Prompt of QA LLM}
\label{qa prompt}
Table \ref{tab:qa_llm} shows the prompt we used for QA LLM. The ``{Wikipedia Docs.}'' will be filled with retrieved Wikipedia documents based on input questions.

\begin{table*}[]
    \centering
    \begin{tabularx}{\textwidth}{X}
    \toprule
    \textbf{Instruction}:\\
         Give a question and some information that may help you answer the question. Please answer the question based on your own knowledge and the information provided. \\
    \hline
    \textbf{Retrieved Information}:\\
    \#\#\# Information\\
    \{Wikipeida Docs.\}\\
    \hline
    \textbf{Input}\\
    \#\#\# Question:\\
    \{Input Question\}\\
    \#\#\# Your Answer:
     (You only need to provide the final answer to the question. Intermediate answers are not needed. Please return your answer in the form of a list, where each element in the list is a short entity answer, such as [Apple]. When you think there are multiple answers, please divide them with a '\#' symbol, such as [Apple\#Banana\#Origin]. If the answer is not included in the information provided, please answer based on your own knowledge. If you don't know either, please return [None].)\\
    \bottomrule
    \end{tabularx}
    \caption{Prompt for QA LLM.}
    \label{tab:qa_llm}
\end{table*}
\begin{table*}[]
    \centering
    \begin{tabularx}{\textwidth}{X}
    \toprule
    \textbf{Instruction}:\\
         Given a subgraph query in the knowledge graph, please transfer it into natural language. The subgraph query is expressed in the format (h,(r,)), where h and r represent the head entity and relation respectively, and the meaning of this query is to find the set of tail entities of h under the relation r. Your responsibility is to transfer it into a question in natural language form. I will give you some examples, please complete your task after reading them:\\
    \hline
    \textbf{Demonstrations}:\\
         \#\#\# Example 1:\\
        Subgraph Query: (Booker T. Jones, (ethnic group,))\\
        Natural Language Question: What is the ethnic group of Booker T. Jones?\\
        \#\#\# Example 2:\\
        Subgraph Query: (Daniel Handler, (educated at,))\\
        Natural Language Question: Where did Daniel Handler receive education?\\
    \hline
    \#\#\# Your Turn (Just output the Natural Language Question and do not return other content):\\
    \textbf{Input}:\\
        Subgraph Query:\\
    \bottomrule
    \end{tabularx}
    \caption{Prompt of verbalization for 1$p$ pattern instances.}
    \label{tab:1p}
\end{table*}
\begin{table*}[]
    \centering
    \begin{tabularx}{\textwidth}{X}
    \toprule
    \textbf{Instruction}:\\
         Given a subgraph query in knowledge graph, please transfer it into natural language. The subgraph query is expressed in the format (h,(r1,r2,)), where h represents the head entity, and r1 and r2 represent a two-hop relation path starting from head entity h. The purpose of this query is to find the target entity associated with the head entity h under the relational path (r1, r2). Your responsibility is to first transfer it into two sub-questions and finally combine them to form a complex question. When constructing the second sub-question, you may need the answer to the first sub-question, so we will assume that the answer to the first sub-question is A1 and the answer to the second sub-question is A2, to facilitate the formulation of the sub-question. When composing the final question, please pay attention to the fluency of the language and avoid mechanically stitching sub-questions together. I will give you some examples, please complete your task after reading them:\\
    \hline
    \textbf{Demonstrations}:\\
        \#\#\# Example 1:\\
         Subgraph Query:(Chongqing, (twinned administrative body, country of citizenship))\\
        Q1: Which city or administrative body that is twinned with Chongqing?\\
        Q1\_Answer: A1\\
        Q2: What is the country of \{A1\}?\\
        Q2\_Answer: A2\\
        Final Question: Which country has a city or administrative body that is twinned with Chongqing?\\
        \#\#\# Example 2:\\
        Subgraph Query:(Inkheart, (cast member, educated at))\\
        Q1: Who is the cast member of Inkheart?\\
        Q1\_Answer: A1\\
        Q2: Where did \{A1\} receive education?\\
        Q2\_Answer: A2\\
        Final Question: Where did the cast member of Inkheart receive education?\\
    \hline
    \textbf{Input}:\\
    \#\#\# Your Turn (Just complete your task in the above format and do not return other content):\\
        Subgraph Query:\\
    \bottomrule
    \end{tabularx}
    \caption{Prompt of verbalization for 2$p$ pattern instances.}
    \label{tab:2p}
\end{table*}
\begin{table*}[]
    \centering
    \begin{tabularx}{\textwidth}{X}
    \toprule
    \textbf{Instruction}:\\
         Given a subgraph query in knowledge graph, please transfer it into natural language. The subgraph query is expressed in the format (h,(r1,r2,r3,)), where h represents the head entity, and (r1,r2,r3,) represents a three-hop relation path starting from the head entity h. The purpose of this query is to find the target entity associated with the head entity h under the relational path (r1,r2,r3,). Your responsibility is to first transfer it into three sub-questions and finally combine them to form a complex question. When constructing the second and third sub-question, you may need the answer to the previous sub-question, so we will assume that the answers to these three sub-questions are A1, A2, and A3, to facilitate the formulation of the sub-question. When composing the final question, please pay attention to the fluency of the language and avoid mechanically stitching sub-questions together. I will give you some examples, please complete your task after reading them:\\
    \hline
    \textbf{Demonstrations}:\\
        \#\#\# Example 1:\\
         Subgraph Query:(Chongqing, (twinned administrative body, country of citizenship))\\
        Subgraph Query: (Android, (developer, country, foundational text))\\
        Q1: Who is the developer of Android?\\
        Q1\_Answer: A1\\
        Q2: What is the country of \{A1\}?\\
        Q2\_Answer: A2\\
        Q3: What is the foundational text of country \{A2\}?\\
        Q3\_Answer: A3\\
        Final Question: What is the foundational text of the Android developer's country?\\
        \#\#\# Example 2:\\
        Subgraph Query: (X-Men: The Last Stand, (cast member, place of birth, shares border with))\\
        Q1: Who is the cast member of X-Men: The Last Stand?\\
        Q1\_Answer: A1\\
        Q2: What is the birthplace of \{A1\}?\\
        Q2\_Answer: A2\\
        Q3: Which area borders with \{A2\}?\\
        Q3\_Answer: A3\\
        Final Question: Which area borders the birthplace of X-Men: The Last Stand's cast member?\\
    \hline
    \textbf{Input}:\\
    \#\#\# Your Turn (Just complete your task in the above format and do not return other content):\\
        Subgraph Query:\\
    \bottomrule
    \end{tabularx}
    \caption{Prompt of verbalization for 3$p$ pattern instances.}
    \label{tab:3p}
\end{table*}
\begin{table*}[]
    \centering
    \begin{tabularx}{\textwidth}{X}
    \toprule
    \textbf{Instruction}:\\
         Given a subgraph query in knowledge graph, please transfer it into natural language. The subgraph query is expressed in the format ``(h1,(r1,)) Intersection (h2,(r2,))'', where h1 and h2 represent two head entities, r1 and r2 are their corresponding relations. The purpose of this query is to find the intersection set of the tail entities of (h1,(r1,)) and (h2,(r2,)). Your responsibility is to first transfer it into two sub-questions and finally combine them to form a complex question. When composing the final question, please pay attention to the fluency of the language and avoid mechanically stitching sub-questions together. The questioning method can be adjusted appropriately, but the meaning cannot be changed. I will give you some examples, please complete your task after reading them:\\
    \hline
    \textbf{Demonstrations}:\\
        \#\#\# Example 1:\\
         Subgraph Query: (Jimmy Carter, (educated at,)) Intersection (John Wells, (educated at,))\\
        Q1: Where did Jimmy Carter receive education?\\
        Q2: Where did John Wells receive education?\\
        Final Question: Where did both Jimmy Carter and John Wells receive education?\\
        \#\#\# Example 2:\\
        Subgraph Query: (Burlington County, (shares border with,)) Intersection (Trumbull County, (shares border with,))\\
        Q1: Which areas border with Burlington County?\\
        Q2: Which areas border with Trumbull County?\\
        Final Question: Which areas border with Burlington County and Trumbull County at the same time?\\
    \hline
    \textbf{Input}:\\
        \#\#\# Your Turn (Just complete your task in the above format and do not return other content):\\
        Subgraph Query:\\
    \bottomrule
    \end{tabularx}
    \caption{Prompt of verbalization for 2$i$ pattern instances.}
    \label{tab:2i}
\end{table*}
\begin{table*}[]
    \centering
    \begin{tabularx}{\textwidth}{X}
    \toprule
    \textbf{Instruction}:\\
         Given a subgraph query in knowledge graph, please transfer it into natural language. The subgraph query is expressed in the format ``(h1,(r1,)) Intersection (h2,(r2,)) Intersection (h3,(r3,))'', where h1, h2 and h3 represent three head entities, r1, r2 and r3 are their corresponding relations. The purpose of this query is to find the intersection set of the tail entities of (h1,(r1,)), (h2,(r2,)) and (h3,(r3,)). Your responsibility is to first transfer it into three sub-questions and finally combine them to form a complex question. When composing the final question, please pay attention to the fluency of the language and avoid mechanically stitching sub-questions together. The questioning method can be adjusted appropriately, but the meaning cannot be changed. I will give you some examples, please complete your task after reading them:\\
    \hline
    \textbf{Demonstrations}:\\
        \#\#\# Example 1:\\
         Subgraph Query: ((Alice in Wonderland, (genre,)) Intersection (Blues Brothers 2000, (genre,)) Intersection (Pinocchio, (genre,)))\\
        Q1: What are the genre of Alice in Wonderland?\\
        Q2: What are the genre of Blues Brothers 2000?\\
        Q3: What are the genre of Pinocchio?\\
        Final Question: What are the same genre shared between Alice in Wonderland, Blues Brothers 2000 and Pinocchio?\\
        \#\#\# Example 2:\\
        Subgraph Query:(Springfield, (capital of,)) Intersection (Ulster County, (shares border with,)) Intersection (Montgomery County, (shares border with,))\\
        Q1: What is the capital of Springfield?\\
        Q2: Which areas border with Ulster County?\\
        Q3: Which areas border with Montgomery County?\\
        Final Question: Which area is the capital of Springfield and borders with Ulster County and Montgomery County at the same time?\\
    \hline
    \textbf{Input}:\\
    \#\#\# Your Turn (Just complete your task in the above format and do not return other content):\\
        Subgraph Query:\\
    \bottomrule
    \end{tabularx}
    \caption{Prompt of verbalization for 3$i$ pattern instances.}
    \label{tab:3i}
\end{table*}
\begin{table*}[]
    \centering
    \begin{tabularx}{\textwidth}{X}
    \toprule
    \textbf{Instruction}:\\
         Given a subgraph query in knowledge graph, please transfer it into natural language. The subgraph query is expressed in the format ``(h1,(r1,)) Intersection (h2,(r2,)) Projection r3'', where h1 and h2 represent two head entities, r1 and r2 are their corresponding relations. The purpose of this query is to fisrt get the intersection set of the tail entities of (h1,(r1,)) and (h2,(r2,)), and then find the tail entities of every entity in the previous intersection set under relation r3. Your responsibility is to first transfer it into three sub-questions and finally combine them to form a complex question. When composing the final question, please pay attention to the fluency of the language and avoid mechanically stitching sub-questions together. The questioning method can be adjusted appropriately, but the meaning cannot be changed. I will give you some examples, please complete your task after reading them:\\
    \hline
    \textbf{Demonstrations}:\\
        \#\#\# Example 1:\\
        Subgraph Query: (John Williams, (educated at,)) Intersection (John Milton, (educated at,)) Projection named after\\
        Q1: Where did John Williams receive education?\\
        Q2: Where did John Milton receive education?\\
        Intersection\_Answer: Inter\_A\\
        Q3: The \{Inter\_A\} was named after what?\\
        Final Question: The place where John Williams and John Milton both received education was named after what?\\
        \#\#\# Example 2:\\
        Subgraph Query: (The Blues Brothers, (cast member,)) Intersection (Going My Way, (cast member,)) Projection member of political party\\
        Q1: Who are the cast members of The Blues Brothers?\\
        Q2: Who are the cast members of Going My Way?\\
        Intersection\_Answer: Inter\_A\\
        Q3: What are the political party of \{Inter\_A\}?\\
        Final Question: What are the political party of people who are cast members of both The Blues Brothers and Going My Way?\\
    \hline
    \textbf{Input}:\\
    \#\#\# Your Turn (Just complete your task in the above format and do not return other content):\\
        Subgraph Query:\\
    \bottomrule
    \end{tabularx}
    \caption{Prompt of verbalization for $ip$ pattern instances.}
    \label{tab:ip}
\end{table*}
\begin{table*}[]
    \centering
    \begin{tabularx}{\textwidth}{X}
    \toprule
    \textbf{Instruction}:\\
         Given a subgraph query in knowledge graph, please transfer it into natural language. The subgraph query is expressed in the format ``(h1,(r1,r2,)) Intersection (h2,(r3,))'', where (h1,(r1,r2,)) represents a two-hop relational path starts from head entity h1 followed by relation r1 and r2, and (h2, (r3,)) is an one-hop relational path start from head entity h2. The purpose of this query is to find the intersection set of the tail entity of relational path (h1,(r1,r2)) and (h2,(r3,)). Your responsibility is to first transfer it into three sub-questions and finally combine them to form a complex question. When constructing second or third sub-questions, you may need the answer to the previous sub-question, so we will assume that the answer to the first sub-question is A1 and the answer to the second sub-question is A2, to facilitate the formulation of the sub-question. When composing the final question, please pay attention to the fluency of the language and avoid mechanically stitching sub-questions together. The questioning method can be adjusted appropriately, but the meaning cannot be changed. I will give you some examples, please complete your task after reading them:\\
    \hline
    \textbf{Demonstrations}:\\
        \#\#\# Example 1:\\
        Subgraph Query: (Drake Bell, (place of birth, shares border with)) Intersection (Santa Ana, shares border with)\\
        Q1: What is the birthplace of Drake Bell?\\
        Q1\_Answer: A1\\
        Q2: Which areas border with \{A1\}?\\
        Q2\_Answer: A2\\
        Q3: Which areas border with Santa Ana?\\
        Q3\_Answer: A3\\
        Final Question: Which regions border Drake Bell's birthplace and Santa Ana at the same time?\\
        Final Answer: A2 Intersection A3\\
        \#\#\# Example 2:\\
        Subgraph Query: (Fran Walsh, (spouse, sport)) Intersection (Fluminense F.C., (sport,))\\
        Q1: Who is the spouse of Fran Walsh?\\
        Q1\_Answer: A1\\
        Q2: What sports does \{A1\} play?\\
        Q2\_Answer: A2\\
        Q3: What sports does Fluminense F.C. play?\\
        Q3\_Answer: A3\\
        Final Question: What sports have Fluminense F.C. and Fran Walsh's spouse played in?\\
        Final Answer: A3 Intersection A2\\

    \hline
    \textbf{Input}:\\
    \#\#\# Your Turn (Just complete your task in the above format and do not return other content):\\
        Subgraph Query:\\
    \bottomrule
    \end{tabularx}
    \caption{Prompt of verbalization for $pi$ pattern instances.}
    \label{tab:pi}
\end{table*}
\begin{table*}[]
    \centering
    \begin{tabularx}{\textwidth}{X}
    \toprule
    \textbf{Instruction}:\\
         Given a subgraph query in knowledge graph, please transfer it into natural language. The subgraph query is expressed in the format ``(h1,(r1,)) Union (h2,(r2,))'', where h1 and h2 represent two head entities, r1 and r2 are their corresponding relations. The purpose of this query is to find the Union set of the tail entities of (h1,(r1,)) and (h2,(r2,)). Your responsibility is to first transfer it into two sub-questions and finally combine them to form a complex question. When composing the final question, please pay attention to the fluency of the language and avoid mechanically stitching sub-questions together. The questioning method can be adjusted appropriately, but the meaning cannot be changed. I will give you some examples, please complete your task after reading them:\\
    \hline
    \textbf{Demonstrations}:\\
        \#\#\# Example 1:\\
         Subgraph Query: (Wuthering Heights, (cast member,)) Union (Traffic, (cast member,))\\
        Q1: Who are the cast members of Wuthering Heights?\\
        Q2: Who are the cast members of Traffic?\\
        Final Question: Who are all the cast members from Wuthering Heights combined with the cast members from Traffic?\\
        \#\#\# Example 2:\\
        Subgraph Query: (Eve, (director,)) Union (Cold Mountain, (cast member,))\\
        Q1: Who is the director of Eve?\\
        Q2: Who are the cast members of Cold Mountain?\\
        Final Question: Please list the director of Eve as well as all the cast members from Cold Mountain.\\
    \hline
    \textbf{Input}:\\
    \#\#\# Your Turn (Just complete your task in the above format and do not return other content):\\
        Subgraph Query:\\
    \bottomrule
    \end{tabularx}
    \caption{Prompt of verbalization for 2$u$ pattern instances.}
    \label{tab:2u}
\end{table*}
\begin{table*}[]
    \centering
    \begin{tabularx}{\textwidth}{X}
    \toprule
    \textbf{Instruction}:\\
         Given two triples with numerical tail entities, please create a comparison-type question based on the given triples and create the corresponding sub-questions for each triple. Finally, you should also give the answer based on the given triple. The final answer should be ``Yes'' or ``No''. Here are some examples:\\
    \hline
    \textbf{Demonstrations}:\\
        \#\#\# Example 1:\\
         Triple 1:(Vietnam male, marriageable age, 20 years old)\\
        Triple 2:(Vietnam female, marriageable age, 18 years old)\\
        Q1: What is the marriageable age for Vietnamese men?\\
        Q2: What is the marriageable age for Vietnamese women\\
        Final Question: Is the marriageable age the same for men and women in Vietnam?
        Answer: No\\
        \#\#\# Example 2:\\
        Triple1:(Vietnam, population, 94660000)\\
        Triple2:(Halifax, population, 424931)\\
        Q1: What is the population of Vietnam?\\
        Q2: What is the population of Halifax?\\
        Final Question: Which company has less population, Vietnam or Halifax?\\
        Answer: Halifax\\
    \hline
    \textbf{Input}:\\
    \#\#\# Your Turn (Just complete your task in the above format and do not return other content):\\
        Subgraph Query:\\
    \bottomrule
    \end{tabularx}
    \caption{Prompt of verbalization for compare pattern instances.}
    \label{tab:compare}
\end{table*}

\end{document}